\documentclass[conference]{IEEEtran}
\IEEEoverridecommandlockouts
\usepackage{cite}
\usepackage{amsmath,amssymb,amsfonts}
\usepackage{url}
\usepackage{algorithmic}
\usepackage{graphicx}
\usepackage{textcomp}
\usepackage{xcolor}
\usepackage{subfigure}
\usepackage{wrapfig}
\usepackage{listings}
\begin{document}

\title{Agent as Cerebrum, Controller as Cerebellum: Implementing an Embodied LMM-based Agent on Drones 

}


\author{\textbf{Haoran Zhao$^{1,3*}$, Fengxing Pan$^{1,3}$, Huqiuyue Ping$^{2,3}$, Yaoming Zhou$^{1*}$}\thanks{$^*$ Corresponding author.}\\
$^1$Beihang University, $^2$Zhejiang University, 
$^3$qingniaoAI 
\\
$^1\{zhaohaoran,panfengxing,zhouyaoming\}@buaa.edu.cn, ^2pinghqy@126.com$
}

\maketitle

\begin{abstract}
In this study, we present a novel paradigm for industrial robotic embodied agents, encapsulating an \verb|agent as cerebrum, controller as cerebellum| architecture. Our approach harnesses the power of Large Multimodal Models (LMMs) within an agent framework known as AeroAgent, tailored for drone technology in industrial settings. To facilitate seamless integration with robotic systems, we introduce ROSchain, a bespoke linkage framework connecting LMM-based agents to the Robot Operating System (ROS). We report findings from extensive empirical research, including simulated experiments on the Airgen and real-world case study, particularly in individual search and rescue operations. The results demonstrate AeroAgent's superior performance in comparison to existing Deep Reinforcement Learning (DRL)-based agents, highlighting the advantages of the embodied LMM in complex, real-world scenarios.

\end{abstract}


\begin{figure*}[t]
    \centering    
    \includegraphics[width=\linewidth]{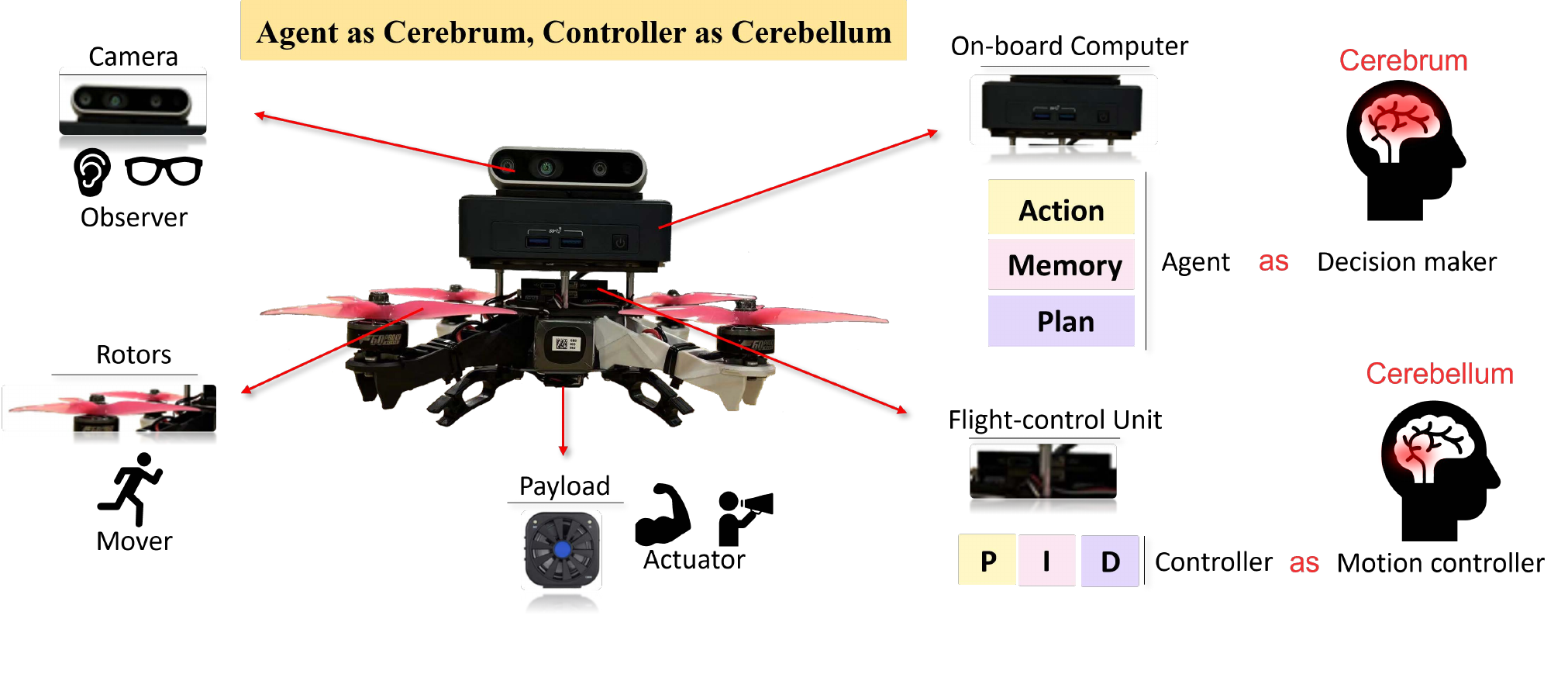}
    \caption{A significant contribution of our work is the proposed architecture depicted herein. This framework guides the integration of embodied agents by leveraging established robot controllers, thereby circumventing the need for agent-based control systems. Accordingly, the agent's role is confined to orchestrating high-level tasks.}
    \label{fig:as}
\end{figure*}

\begin{figure*}[t]
    \centering
    \includegraphics[width=\linewidth]{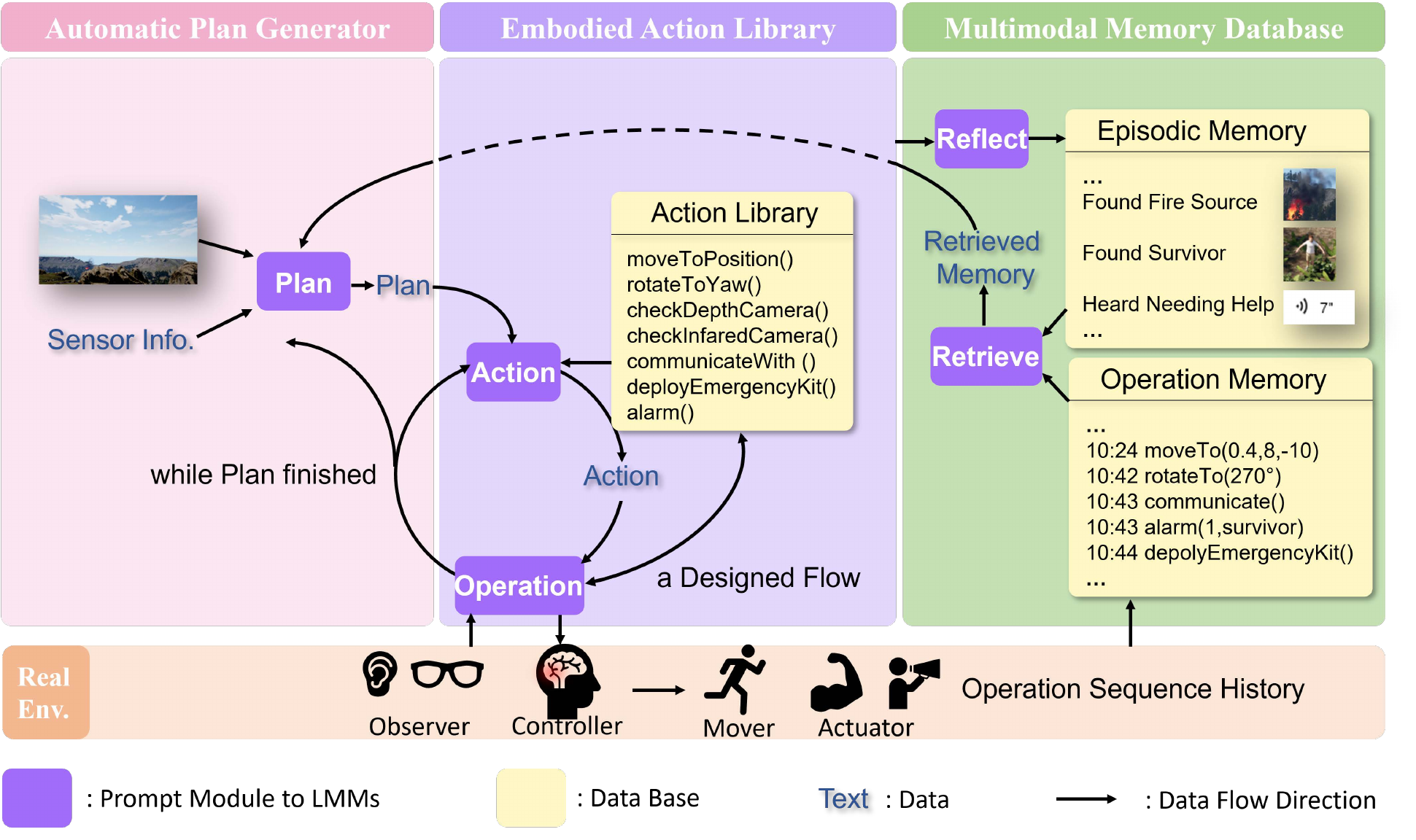}
    \caption{The Architecture of AeroAgent. This figure illustrates the AeroAgent architecture, which encompasses an embodied plan generator, a multimodal memory, and an integrated action library. It also delineates the task execution workflow within AeroAgent.}
    \label{fig:arch}
\end{figure*}

\section{Introduction}
In end-to-end robot learning, tasks are typically acquired through imitation learning or reinforcement learning strategies, requiring the accumulation of task-specific datasets. These datasets may focus exclusively on singular tasks or span multiple tasks to enhance the robot's capabilities in complex environments\cite{DBLP:journals/corr/abs-1806-10293}\cite{zhang2018deep}\cite{kalashnikov2021mtopt}. This methodology aligns with traditional supervised learning practices observed in fields such as computer vision and natural language processing (NLP), where the creation and utilization of task-specific datasets are standard to solve distinct challenges. However, this paradigm encounters several limitations:
(1) Large-scale annotated data is a fundamental requirement for end-to-end training, which may not be feasible or practical in many scenarios.
(2) The ability to generalize knowledge and skills to new tasks or different robotic platforms remains largely inefficient.
(3) The simulation-to-reality transition poses considerable challenges due to discrepancies between simulated and real-world environments.


In recent years, a transformative shift has occurred in fields such as computer vision and natural language processing. This shift has moved the focus away from isolated, small-scale datasets and models toward expansive, universal models that have been pretrained on large datasets. The key elements underlying the success of these models include their ability to leverage open-ended, task-agnostic training and their employment of high-capacity neural architectures, which are designed to integrate the extensive knowledge contained within these datasets.
Pretrained high-capacity models, rooted in comprehensive web-sized datasets, have emerged as a robust and efficient foundation for a multitude of subsequent tasks. For example, advanced language models have demonstrated capabilities extending beyond generating fluent text, as evidenced by references\cite{anil2023palm}\cite{openai2023gpt4}, to facilitating emergent problem-solving\cite{cobbe2021training} and programming code generation\cite{li2023large}. Additionally, robotics applications have benefited from this paradigm by utilizing large models to enhance generalization, multitasking, and real-world operational abilities, such as those seen in robot transformer frameworks \cite{brohan2023rt1}\cite{brohan2023rt2}, embodied large models \cite{anil2023palm}, and robot manipulation models \cite{bharadhwaj2023roboagent}.

Advancements in robotic manipulation have markedly progressed, especially for large-scale models, yet notable challenges persist in designing robots tailored for specific tasks such as quadruped robots or drones. (1) These high-capacity models often exhibit slow responsiveness, which may fail to satisfy the precise and swift control requisites of certain robotic applications. For instance, while a stationary robotic manipulator arm may successfully execute operations within a 2-4 second timeframe, a drone necessitating continual stability must achieve control at intervals as brief as 0.01 seconds. (2) Systems leveraging these models typically exhibit a lack of redundancy to efficiently cope with unforeseen circumstances that might arise during operation. (3) The embodiment of intelligence in these task-driven robots—and consequently, their operational efficiency—remains suboptimal, necessitating further research and development to bridge these gaps.


With the advancement of large pre-trained model techniques, numerous recent studies have leveraged these expansive models, typified by GPT-4\cite{openai2023gpt4}, as foundation models\cite{bommasani2022opportunities}. Indeed, LLM(Large Language Model)-based agents\cite{xi2023rise}, hereafter referred to as LLM-powered autonomous agents\cite{weng2023AutoAgents}, have exhibited remarkable proficiency in addressing multifaceted problems across various domains. These include multimodal content generation\cite{shen2023hugginggpt}, software development\cite{qian2023chatdev}, social simulations\cite{park2023generative}, and task-oriented evaluative scenarios\cite{lin2023agentsims}\cite{liu2023agentbench}. Within an LLM-based agent framework, the LLM serves as the cognitive epicenter, synergizing with essential components such as planning, memory, and tool use\cite{weng2023AutoAgents}.

The tool use component grants the agent the facility to leverage auxiliary tool modules, thereby permitting the agent to channel its efforts into crafting high-level strategies, while delegating the execution of subordinate, low-level operations to specialized modules. For example, within the context of drones, if we conceptualize the path planning and control systems as a unit tasked with generating commands, the agent can then delineate intended waypoints without the onus of formulating granular operational directives. The memory module proficiently assesses evolving scenarios, thus equipping the agent with the capacity to engage in few-shot learning when confronted with unfamiliar assignments. Furthermore, the planning module adeptly deconstructs objectives by employing systematic reasoning, and through introspection, it reviews the tools at its disposal, effectively harnessing embodied intelligence capabilities.

Large Multimodal Models (LMMs) extend upon the capabilities of LLMs by integrating multi-sensory processing abilities\cite{yang2023dawn}. By harnessing world knowledge through the analysis of diverse multimodal data, such as visual, auditory, and textual information, LMMs are exceptionally equipped for deployment in robotic systems that interact with and respond to a wide range of sensory inputs. To this end, we have developed a robotic agent that leverages the sophisticated faculties of LMMs. Further, we introduce an framework designed for LMM-based agents implementing on drones, aimed at enhancing the efficiency and performance of various industrial applications. Our empirical research includes a series of experiments within simulated environments as well as practical demonstrations, each underscoring the efficacy and potential of this cutting-edge technology in real-world settings.

The principal contributions of this work are enumerated below:

\begin{itemize}
    \item We introduce a novel paradigm that conceptualizes \verb|agent as cerebrum , controller as| \verb|cerebellum| for the execution of industrial tasks, as depicted in Fig. \ref{fig:as}. This framework has been concretely realized in industrial drones through both simulated experiments involving 4 tasks and a real-world case study on individual search and rescue, thereby bridging a noticeable gap in existing research.
    \item We present an agent architecture premised on LMM approaches. In contrast with prevalent LLM-based agents, our architecture exhibits enhanced aptness for robotic interactions within the physical environment and leverages embodied intelligence more effectively.
    \item We unveil ROSchain, a framework designed to enable the development of robotic applications that harness the capabilities of both LLMs and LMMs. ROSchain streamlines the integration of large-scale models with a robot's sensory apparatus, execution units, and control mechanisms through a suite of modules and Application Programming Interfaces (APIs).

\end{itemize}





\section{Preliminaries}
\subsection{Agent}

Within the realms of artificial intelligence and automation, an agent is broadly defined as an autonomous entity capable of perceiving—also referred to as observing—its environment and executing actions aimed at achieving specific goals. The decisions made by an agent are rooted in its ability to autonomously evaluate its surroundings \cite{xi2023rise}.

\textbf{DRL-Based Agent.} At the core of Reinforcement Learning (RL) lies the goal of formulating a policy, $\pi(\mathbf{a}_{t} | \mathbf{s}_{t})$, which dictates an action distribution contingent upon the states. This policy engenders the RL-based agent with decision-making faculties. Deep Reinforcement Learning (DRL) augments RL agents' perceptual modalities to include multimodal image observations, thereby increasing their applicability in various experimental robotics scenarios\cite{zhou2023train}.

\textbf{LM-Based Agent.} The Large models (LMs) mainly refer to LLMs and LMMs in this section. Since there are not yet any other work on LMM-based agents, some basic concepts about LLM-based agents are introduced. The LLM-based agent operates as a low-code, autonomous workflow utilizing natural language as its primary data interface. Large Language Models serve as the central brain or decision-making system\cite{xi2023rise}. These agents enhance their perception and action repertoires by incorporating advanced techniques like multimodal perception and effective tool use. Capable of demonstrating both reasoning and planning skills, LLM-based agents utilize methods such as Chain-of-Thought (CoT) and problem decomposition. Memory, regarded here as the mechanism for acquiring, preserving, retrieving and utilizing information, equips the agent with the ability to engage in few-shot learning; this involves rapidly adapting to new tasks based on minimal exposure to new data, propelled by dynamic feedback and novel action execution.

\subsection{Controller}
The controller within an aircraft refers to its flight control system, which manages the rudders based on sensor data. This system adheres to programmed logic to ensure the aircraft maintains a predetermined flight behavior, navigational path, or heads towards a specific waypoint. For multirotor aircraft, it achieves this by manipulating the rotor speeds to control flight, hover, and attitude adjustments. Utilizing sensor-provided attitude information, the system processes this data, assesses the current state, and commands actuators to modify the aircraft's orientation accordingly.

Control laws, embedded in the flight control system, dictate the generated control commands. These laws define the relationship between controlled state variables and input signals. The development of control algorithms is aimed at achieving a closed-loop system that is stable, responsive, and suffers minimal overshoot. Central to control theory, these algorithms adjust controller parameters to satisfy the closed-loop transfer function's requirements.

\begin{figure*}[t]
    \centering    \includegraphics[width=0.9\linewidth]{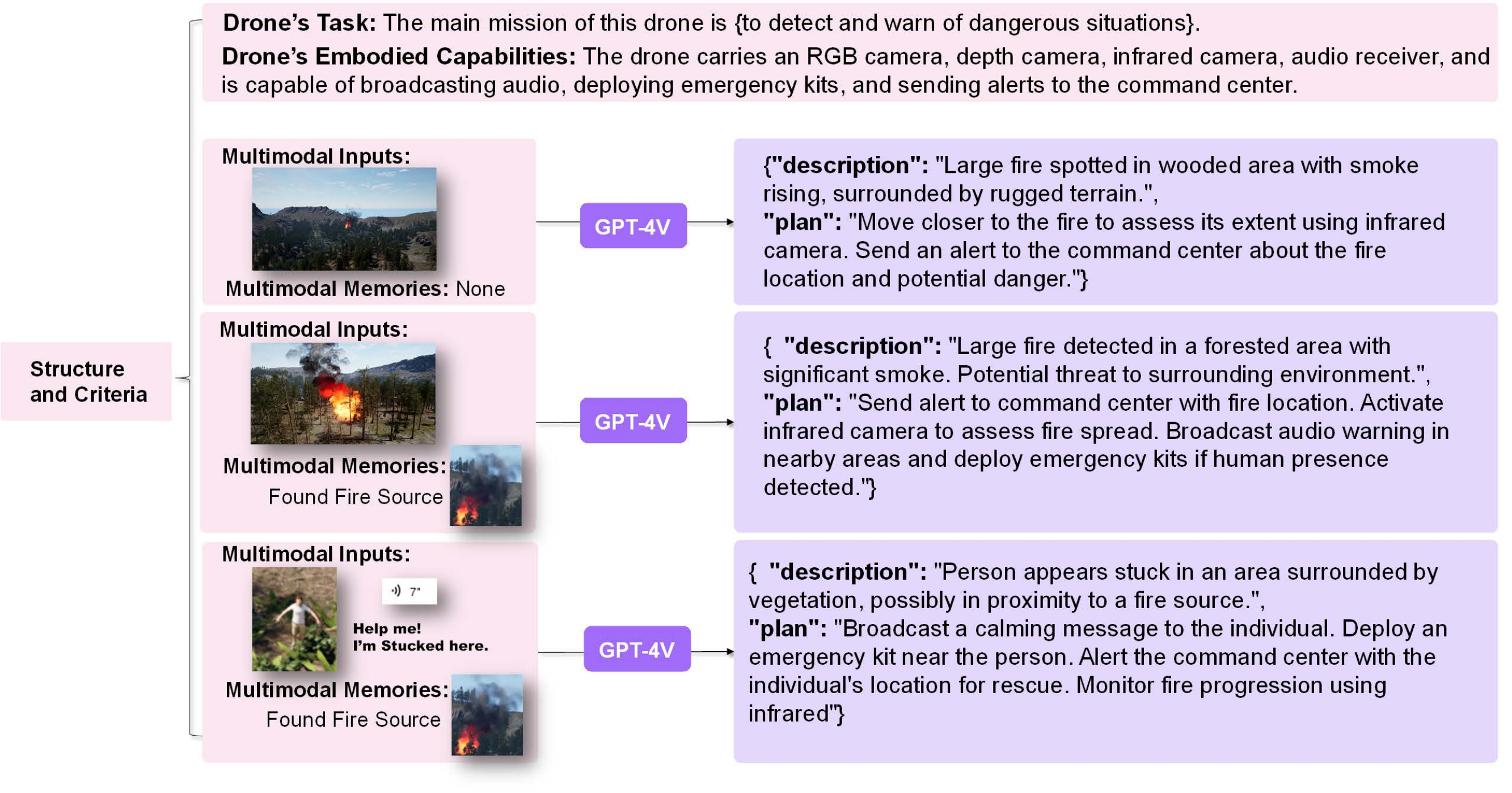}
    \caption{Examples of the Automatic Plan Generator. Some structure and criteria prompts are not listed. It can generate descriptions and contingency plans for multimodal data.
}
    \label{fig:plan}
\end{figure*}

\section{AeroAgent: an Autonomous Embodied Agent with Large Multimodal Models}
AeroAgent is designed as an autonomous embodied agent which can perceive, reasoning and take actions to complete multiple unplanned tasks 
rather than a specific task. 
In this section, we will introduce the key components of the agent which equipped the agent with decision-making capabilities. Then, we will show and evaluate these capabilities. We will introduce drone task as example.
\subsection{Components}
AeroAgent consists of three main components: (1)an automatic plan generator (Sec. \ref{planning}) with multimodal perceptual monitoring,(2)a multimodal memory database (Sec. \ref{memory}) for retrieval and reflection and (3)an embodied action library (Sec. \ref{action}) for stable execution. Fig.~\ref{fig:arch} illustrates the architecture of AeroAgent. 

\subsubsection{Automatic Plan Generator}\label{planning}
The deployment of robots, notably drones, across a multitude of environments necessitates a level of autonomy that allows them to perform varied tasks independently. Traditional methods, such as behavior trees or finite state machines, falter when defining precise workflows in advance due to the unpredictability of these settings. Consequently, it is imperative for robots to possess the capability to dynamically generate plans that take into account the real-time context of their surroundings. Currently, a prevalent practice involves leveraging robots as distant proxies of human operators, who control these machines using advanced sensory systems. 

Nevertheless, this methodology is heavily dependent on human intervention and expertise, limiting the robots' potential for genuine autonomous decision-making. The imminent move in the field is towards fostering intelligent robots capable of independently orchestrating their actions in response to the complexities of their operational environment. Achieving this will lead to substantial advancements in autonomous multi-task coordination under uncertain conditions. Such mastery calls for an amalgamation of skills—including perception, learning, planning, and control—enabling the robots to devise and fine-tune strategies based on live environmental feedback. Meeting this demand represents one of the significant scientific challenges presently confronting robotics research.

In this work, we present the design of an Automatic Plan Generator, as depicted in Fig. \ref{fig:plan}, which incorporates LMMs to effectively monitor and process multimodal situational inputs. This system is capable of autonomously crafting open-ended plans, allowing for the generation of a wide array of task scenarios that are essential to the sophisticated decision-making capabilities of autonomous agents.

The input to LMMs consists of several components:

(1) Structured prompts are used to guide and constrain the plan generation;

(2) Drone's task described as natural language, such as \verb|"to detect and warn of dangerous situations"|;

(3) Drone's embodied capabilities;

(4) Multimodal environment observations, including images, audio and location, perceived by various sensors;

(5) Memories retrieved from the memory database, including episodic memories and operation sequence memories.

\subsubsection{Multimodal Memory Database}\label{memory}

\begin{figure*}[t]
    \centering    \includegraphics[width=\linewidth]{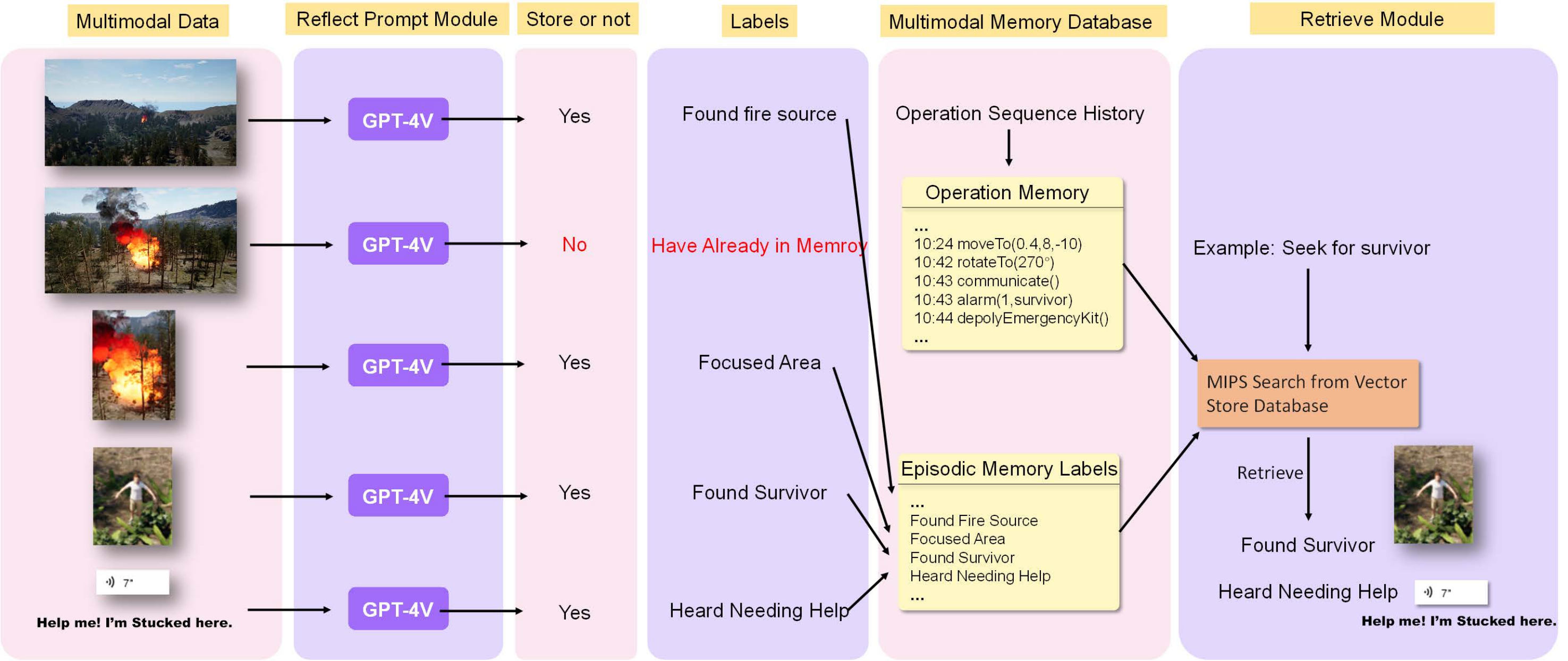}
    \caption{Examples of the multimodal memory database and its reflection and retrieval. Taking seek for survivor as an example, memory reflecting and retrieving are implemented based on labels.}
    \label{fig:mem}
\end{figure*}

The AeroAgent is endowed with few-shot learning abilities, courtesy of its memory module, which extends the capabilities of large-scale multimodal models. By harnessing the sophisticated reasoning prowess of such models, the memory feature empowers agents to assimilate knowledge progressively. LLM-based agents primarily exploit vector databases for the efficient reflecting and retrieving of memories, thus enabling the fetching of pertinent memories from extensive repositories by amalgamating similarity searches with additional parameters.

Contemporary research \cite{gershman2017reinforcement}\cite{lengyel2007hippocampal}\cite{bornstein2017reminders}\cite{bornstein2017reinstated} has increasingly underscored the utility of episodic memory in animals and humans over semantic memory. These episodic memories are intrinsically encoded in the brain in the form of images and sounds -- a modality that presents formidable challenges for direct database-like retrieval.

In light of these challenges, we have conceived a memory tagging strategy that simplifies the reflection and retrieval of episodic multimodal memories. This technique entails encoding multimodal experiences and their narratives into the memory database whenever they contain salient information. Consequently, segments of multimodal episodic memories—comprising multi-faceted data and their corresponding labels—are cataloged within the multimodal memory database. These labels, alongside other memory types, are stored in a vector database for efficient recall. When retrieving memories, the vector database is scoured; upon locating the label of a multimodal episodic record, both the label and its associated multimodal content are then fed into the LMM for processing.

Additionally, records of historical operations are also stored in memory.
These two parts constitute our multimodal memory database (Fig. 
\ref{fig:mem}).

\subsubsection{Embodied Action Library}\label{action}

Action Libraries are essential for executing embodied intelligence tasks in robotic systems. It is widely acknowledged that large-scale model architectures often struggle with issues of perceptive inconsistencies, commonly referred to as "hallucination problems." Moreover, the varied training datasets for these complex models encompass a broad range of robotic actuators. Without appropriate constraints, there is a significant risk that the model's output may prove incompatible with the actuators' capabilities. While employing few-shot learning mechanisms in conjunction with a memory system, as discussed in \cite{wang2023voyager}, can mitigate this issue somewhat, it is not a panacea. For instance, allowing a drone to navigate and learn from unconstrained exploration in a physical environment may hinder its task efficacy and introduce safety hazards.

To address these challenges, we have engineered specialized action librariy tailored for drones equipped with embodied intelligence. Task specificity is a determining factor for the payload configurations a drone may carry, with each configuration demanding distinct execution modalities. This action library is curated to match specific combinations of payloads, delineating a suite of feasible actions. When an agent decomposes a plan into specific actions, the first step is to consult the action library to determine the appropriate actions for the plan. Subsequently, the selected action activates a predefined actuator workflow to infer, process, and determine necessary parameters, effectively guiding the corresponding actuators.

\begin{figure*}[t]
    \centering    \includegraphics[width=\linewidth]{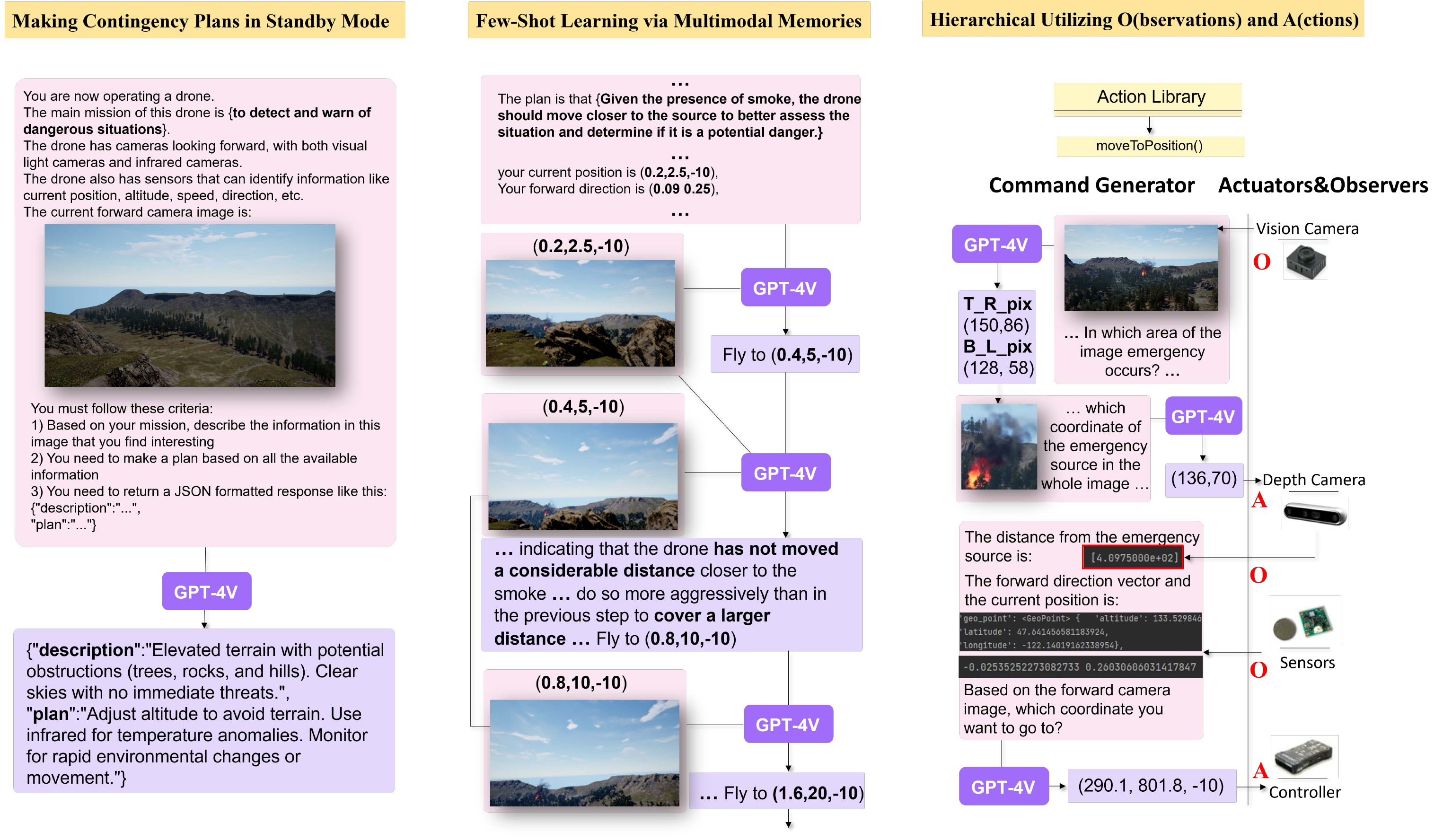}
    \caption{Capabilities Example. Capabilities are Making Contingency Plans while on Standby, Few Shot Learning via Multimodal Memory Hierarchical Utilization Observations and Action from left to right}
    \label{fig:cap}
\end{figure*}

\begin{itemize}
    \item Select an action function from the Action Library. This action function indexes a specific operations flow. At the same time, the parameters of this action function need to be determined. For example, when \verb|moveToPosition()| is selected, there are three undetermined parameters $(x,y,z)$.
    \item Execute this operations flow. This flow contains access to sensor results and manipulations of actuators, as well as actively invoking a particular sensor. After the flow ends, the parameters of the action function will be determined, and then the final action operation will be executed. A specific example is shown in the Fig.\ref{fig:cap}, right. 
\end{itemize}

\subsection{Capabilities}

After the development of such a novel LMM-based embodied agent, we observed that it possesses certain distinctive capabilities, which are not easily quantifiable, in contrast to agents developed with Deep Reinforcement Learning (DRL), which is currently predominant in the field of robotics. The intriguing nature of these capabilities necessitates a qualitative introduction; therefore, we offer illustrative case studies prior to engaging in quantitative comparative experiments\footnote{\footnotesize In this paper, we conduct experiments employing GPT-4V, particularly its gpt-4-vision-preview model (released by OpenAI on November 6th, 2023).}. These capabilities, which emerge naturally from the architecture of AeroAgent, convey its proficiency in handling tasks within practical environments, as depicted in Fig. \ref{fig:cap}. Subsequent sections of this study are dedicated to thorough experimental analyses exploring the empirical performance of AeroAgent, thereby augmenting our initial qualitative assessment with quantitative data. 

\begin{figure*}[t]
    \centering    \includegraphics[width=\linewidth]{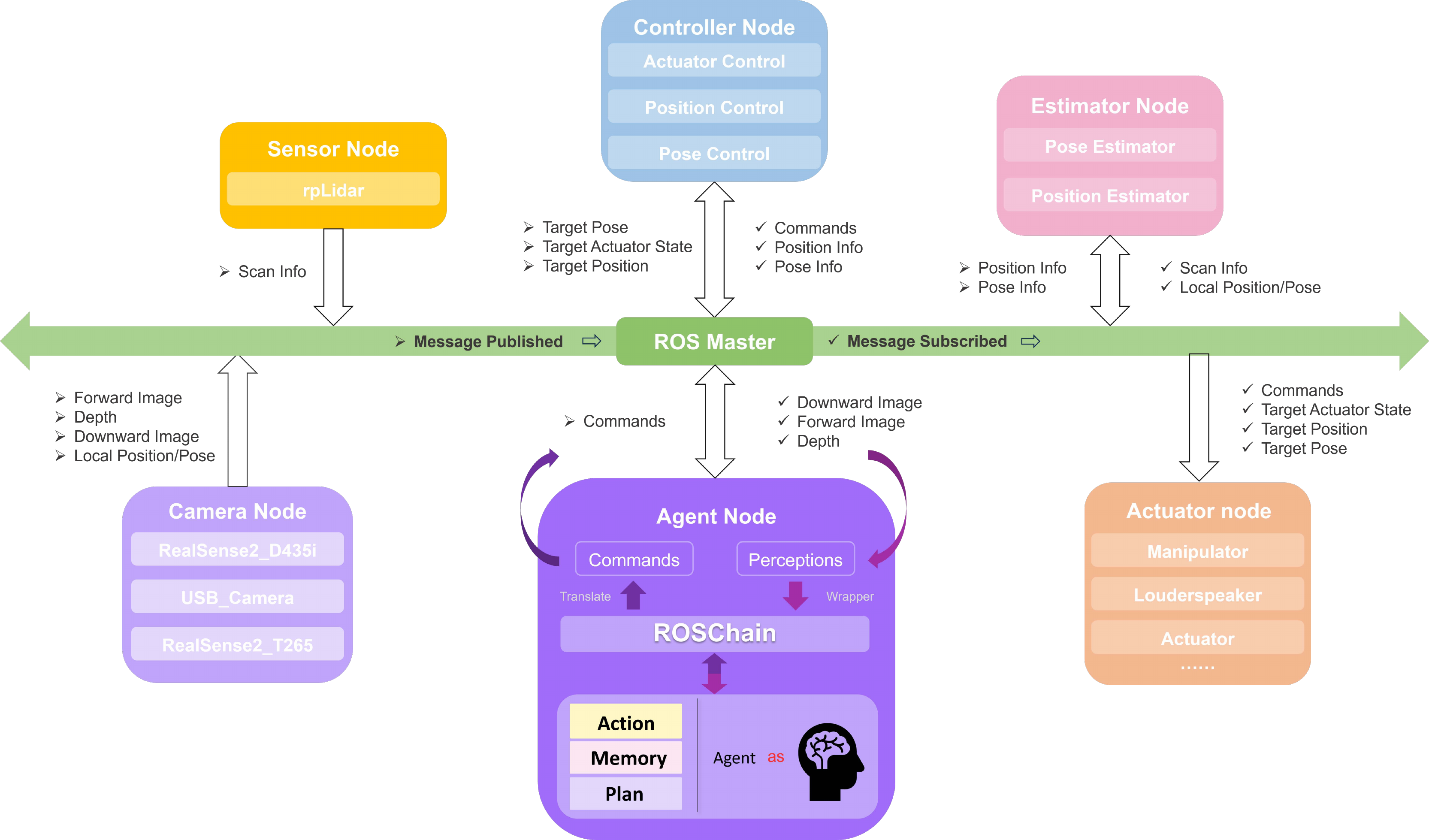}
    \caption{The architecture between nodes in our system. Nodes are classified into six functional categories: Agent Node, Controller Node, Sensor Node, Camera Node, Estimator Node, and Actuator Node. Communication between these nodes—consisting of publish-subscribe messages or request-response services—is coordinated via the ROS Master, which functions analogously to a bus in a computer system, managing data transmission pathways throughout the network.}
    \label{fig:node}
\end{figure*}

\subsubsection{Making Contingency Plans in Standby Mode}

DRL has marked significant successes in domains necessitating rapid decision-making, such as gaming and robotic control. Yet, its efficacy diminishes in scenarios that demand long-term strategic planning—particularly in monitoring and surveillance tasks characterized by infrequent state changes. DRL faces challenges in these settings as it processes environmental information into discrete states for decision-making, rendering its policy functions unable to proactively recognize a distinction between states that represent prolonged stability and those signaling abrupt disruptions. To a DRL-based agent, these scenarios merely represent disparate state configurations meriting responses that maximize immediate rewards—often engendering behaviors that are suboptimal for surveillance tasks, such as unnecessary movements toward states with ostensibly higher value predictions. While some studies attempt to curb this issue by incorporating auxiliary rewards to penalize energy consumption or incentivize stationary surveillance, such approaches can inadvertently promote lazy-learning.

The AeroAgent exhibits the capability of making contingency plans in standby mode.  The AeroAgent integrates multimodal data streams and temporal comparisons across sequential frames to inform its decision-making process, engendering mission-oriented plan formulation. 
The manifestation of such plans is \verb|"Monitor for rapid environmental changes"|, which gets translated into concrete actions of maintaining position or invoking other sensors to acquire more information, rather than the \verb|"|imagined\verb|"| movements of reinforcement learning agents.

\subsubsection{Few-Shot Learning via Multimodal Memories}
Few-shot learning through contextualization is a recognized strength of LMMs and LLMs. Enhancing this capability, our AeroAgent integrates a multimodal memory database to leverage the few-shot learning potential of LMMs more effectively. The AeroAgent is specifically engineered to utilize a vast array of sensory data and records within this database. An illustration of its enhanced capability is AeroAgent's ability to estimate its current location by cross-referencing recently captured camera images in the multimodal memory database with those previously encountered, evaluating the variation in visual details. Furthermore, the AeroAgent intelligently synthesizes temporal and ancillary data to make real-time decisions, such as whether an acceleration towards the documented destination in the memory database is warranted. This strategic integration of multimodal data showcases AeroAgent's superior situational assessment and decision-making skills in dynamic environments.

\subsubsection{Hierarchical Utilizing Observations and Actions}
In contrast to the typically rigid processing pipeline of end-to-end learning systems, the AeroAgent offers a more flexible approach to managing both perceptual and execution layers. In the domain of DRL-based agents, it is common practice to standardize the agent's sensory inputs into a uniform data format before deriving either a set of parameters for actions in a continuous space or electing a discrete action. Conversely, an agent built on an LMM can assimilate environmental inputs of varying sources and semantic nature in a hierarchical manner, mirroring the sequential focus of human attention, where certain stimuli prompt the prioritization of pertinent information. By synthesizing data from disparate sources, the agent possesses the capability to deduce or even proactively seek out supplementary focused observations. Furthermore, an LMM-based agent's action repertoire is not limited to predefined choices or static responses. The act of interrogating a sensor may itself constitute an action, and the determination of action parameters can emerge through logical inference rather than being confined to optimization via policy gradients. It is this sophisticated employment of perception and action coordination that we define as Hierarchical utilization in observing and acting.

\section{Implementation on an Industrial Drone with ROSchain}

In this chapter, we present practical implementation cases of AeroAgent in the realm of industrial drones. These drones are employed across various industrial sectors, often under demanding work conditions and complex tasks that necessitate a high level of autonomy for effective task execution. Specifically, industrial drones have gained prominence in critical applications such as emergency rescue operations, power line and pipeline inspections, precision agriculture, construction site surveillance, and more.

To seamlessly integrate the functionalities provided by AeroAgent into real-world drone systems, we have developed ROSchain, a framework based on the Robot Operating System (ROS)\cite{ros}. ROSchain facilitates communication between large-scale models and the drones' sensory, execution, and control mechanisms through a dedicated set of modules and APIs. Within the ROS, AeroAgent operates as a node that actively publishes and subscribes to messages, enabling dynamic system interactions through the ROSchain adapter.


\subsection{Implementation}
The hardware implementation of the drone is shown in Fig. \ref{fig:as}. 
We leverage the most widely-use ROS to implement the drone's system architecture.
The Robot Operating System is a set of software modules and tools that help one build robot applications. In ROS, \textit{Nodes} are processes that execute computational tasks. \textit{Nodes} communicate with each other by passing \textit{Messages}. \textit{Messages} are conveyed via \textit{Topics} to \textit{publish} and \textit{subscribe}, or via \textit{Services} to \textit{request} and \textit{response}.

Within our ROS-based system architecture, the agent and the controller operate as distinct nodes, each producing different levels of action through message exchanges coordinated by the ROS master. The agent resembles the cerebrum, generating high-level actions and providing overall node management, whereas the controller resembles the cerebellum, executing precise motor control upon receiving movement commands from the cerebrum. Concurrently, when the cerebrum emits a directive for high-level actions, such as operating specific actuators, the corresponding nodes execute a response. Figures detailing the nodes incorporated into our system can be found in Fig. \ref{fig:node}.

\textbf{Agent Node.} This node is pivotal in high-level decision-making and command issuance. It is equipped with ROSchain to wrap sensory information into agent's perceptions and translate the agent's operations into actionable commands.

\textbf{Controller Node.} Tasked with receiving directives from the agent node, this node specializes in executing low-level motion control operations to coordinate the robotic system’s movements.

\textbf{Sensor Node.} This component is integral for relaying sensor data, encompassing devices such as \verb|rpLidar| that contribute to the robot’s awareness of its surroundings.

\textbf{Camera Node.} In conjunction with various imaging sensors, such as \verb|RealSense2_T265_Camera|, \verb|RealSense2_D435i_Camera|, and \verb|USB_Camera|, this node acquires visual data crucial for tasks including navigation, forward surveillance, and vertical reconnaissance, respectively.

\textbf{Estimator Node.} Assigned the crucial role of deducing the robotic system’s spatial orientation and positional coordinates, this node is a cornerstone for precise localization and maneuvering.

\textbf{Actuator Node.} Upon reception of the action commands, this node bears the responsibility for the articulation of the robot’s actuators, enabling the conversion of commands into physical actions.


\subsection{ROSchain}

Large models, as well as agents, require APIs or tools to access necessary resources. A comprehensive framework, or toolkit, can offer integrated solutions as demonstrated by \cite{langchain}. An embodied agent requires a toolkit to enable interaction with the physical environment. Such is the function of ROSchain, which facilitates the translation of operations within an agent into executable commands, and captures sensory data from the physical world, wrapping it in a way that is comprehensible to the agent.

ROSchain comprises several core components, including system initialization, a parameter module, and a function module. It equips agents with the ability to register nodes, subscribe to topics, publish topics, and request services on the ROS. This integration process simplifies the incorporation of agents into existing robotic systems or the development of new systems, adhering to ROS's standardized conventions. The principal invocation methods provided by ROSchain are delineated below:


\begin{lstlisting}
import ROSchain
import agent

roschain=ROSchain("agent")

perception = roschain.subscriber("Camera",format_name="Image")
plan = agent.plan(perception, ...)
action = agent.action(plan, ...)
operation, config = agent.operation(action, ...)

result = roschain.request("active_observation", operation, config, respond_format_name="String")

roschain.publish("command", operation=result["operation"], config=result["config"])
\end{lstlisting}




The main function and capabilities provided by ROSchain are defined as follows: 

\begin{lstlisting}
import rospy
import base64
from std_msgs.msg import String
from sensor_msgs.msg import Image

class ROSchain:
    def __init__(self, node_name):
        rospy.init_node(node_name)
        self.queue_size = parameters["queue_size"])
    def subscriber(self, message_name, format_name):
        sub = rospy.Subscriber(message, String if format_name=="String" else Image, doMsg = self.wrapper,queue_size=parameters["queue_size"])
        rospy.spinonce()
        return sub
    def publish(self, message_name, operation, config, format_name):
        command = self.translate(operation, config)
        pub = rospy.Publisher("message_name", String, self.queue_size
        msg = String(command)
        pub.publish(msg)
    def request(self, service_name, operation, config):
        rospy.wait_for_service(service_name)
        try:
            client = rospy.ServiceProxy(service_name, self.wrapper)
            response = client(self.translate(operation, config))
            return response
        except rospy.ServiceException, e:
            print ("service call failed: %s"%e)
        
    def wrapper(self, message):
        if message.type == "String":
            return message
        elif message.type == "Image":
            return base64.b64encode(message).decode("utf-8")
        ...
        
    def translate(self, operation, config):   
        return self.command_dict[operation][config]
        ...

\end{lstlisting}





Messages that are to be published to the ROS take the form of commands, the scope of which must be delineated with respect to the functional capacity of interconnected ROS nodes. Within our developed implementation example, known as AeroAgent, we deisign three distinct categories of commands: controller command, execution and active observation. Several of these commands necessitate the input of parameters. While a subset of the requisite parameters is pre-initialized in the ROSchain parameter module, others must be dynamically ascertained based on either the published messages or the services' return values from other nodes. The organizational scheme of these commands is systematically detailed in Table~\ref{tab0}.

\begin{table}[h]
\caption{Commands in our implementation}
\begin{center}
\begin{tabular}{|c|c|c|c|}
\hline
\textbf{Command}&\multicolumn{3}{|c|}{\textbf{Command Information}} \\
\cline{2-4} 
\textbf{Type} & \textbf{\textit{Command}}& \textbf{\textit{Publish/Client}}& \textbf{\textit{Subscribe/Server}} \\
\hline
& \textit{Move\_ENU}& \textit{Agent}&\textit{Controller}  \\
\cline{2-4} 
&\textit{Move\_Body} & \textit{Agent}&\textit{Controller}  \\
\cline{2-4} 
& \textit{Takeoff}& \textit{Agent}&\textit{Controller}  \\
\cline{2-4} 
\textbf{Controller} & \textit{Land}& \textit{Agent}&\textit{Controller}  \\
\cline{2-4} 
\textbf{Command}& \textit{Arm}& \textit{Agent}&\textit{Controller}  \\
\cline{2-4} 
& \textit{Disarm}& \textit{Agent}&\textit{Controller}  \\
\cline{2-4} 
& \textit{Failsafe}\_land& \textit{Agent}&\textit{Controller}  \\
\cline{2-4} 
& \textit{Idle}& \textit{Agent}&\textit{Controller}  \\
\hline
\textbf{Execution}& \textit{Loudspeaker}& \textit{Agent }&\textit{Loudspeaker}  \\
\cline{2-4} 
\textbf{Command} & \textit{Manipulator}& \textit{Agent }&\textit{Manipulator}  \\

\hline
& \textit{Depth\_Observe}& \textit{Agent} & \textit{Depth Camera}  \\
\cline{2-4} 
\textbf{Active} & \textit{Infrared\_Observe}& \textit{Agent}&\textit{Infrared Camera}  \\
\cline{2-4} 
\textbf{Observe}& \textit{Lidar\_Observe}& \textit{Agent}&\textit{Lidar}  \\
\cline{2-4} 
& \textit{Down\_Observe}& \textit{Agent}&\textit{Downward Camera}  \\
\hline
\end{tabular}
\label{tab0}
\end{center}
\end{table}



\section{Experiments}

In this chapter, we conduct a comprehensive evaluation of the AeroAgent's performance, contrasting it with established benchmarks. Our comparative analysis begins by reviewing the agent's architecture through a comparison with single call of LMM. Next, we scrutinize the integral role of ROSchain by performing an ablation study to ascertain its impact on our system. Additionally, we juxtapose the AeroAgent with a DRL-based agent, widely recognized for their capability in robotic decision-making processes.

Given that existing evaluation standards tend to concentrate on task-specific accomplishments within simulated settings, and our approach prioritizes multi-tasking in authentic environments, opting out of some RL benchmarks was a deliberate choice. To facilitate a just assessment between AeroAgent and DRL-based agents, we adhere to these key tenets in our experimental evaluations:

(1) We acknowledge that empirical comparisons within physical settings showcase AeroAgent's robustness more distinctly than in simulations since DRL-based agents encounter limitations in applying comprehensive interactive sampling in practical scenarios. Nonetheless, DRL-based agents often integrate Sim2Real strategies to bridge the gap between simulation and reality. By opting for high-reliability simulated environments for our comparison, though we mitigate AeroAgent's inherent advantages, the objectivity and persuasiveness of our findings is reforced.

(2) In an industrial context, agents must be pragmatic in terms of task execution concerning temporal and computational expenditures. DRL-based agents, therefore, must be trained within the realistic confines of limited computational resources and time (e.g., training restricted to one day), bypassing the pursuit of algorithmic perfection through extensive hyperparameter optimization which is not feasible in industrial settings.

(3) Adopting the traditional reinforcement learning paradigm, where we treat everything external to the AeroAgent as the \textit{environment} and the agent's incoming and outgoing messages as \textit{observations} and \textit{actions} respectively, we align the AeroAgent's interaction intervals with those in RL. Specifically, one \textit{step}—the reception of an observation followed by the issuance of an action—serves as a core metric for evaluating algorithmic efficiency. Balancing interaction frequency is vital to prevent undue hardware deterioration and control temporal costs.

In addressing the inherent challenges and intricacies posed by multimodal data, we have employed AirGen\cite{vemprala2023grid}, a simulation platform of high fidelity designed explicitly for aerial robotics scenarios. Provided with the capability to simulate a comprehensive range of environments, both synthetic and geographically-specific real-world settings, AirGen is pivotal for producing an assortment of diverse scenarios. Such variety is essential to rigorously assess the robustness and adaptive capacity of robotic methods. Within AirGen, we have carefully selected a suite of tasks that are critical for our investigation: wildfire search and rescue, vision-based landing, infrastructure inspection and safe navigation. These tasks are strategically chosen for their relevance in testing the efficacy of robotics algorithms in dynamic and challenging real-world situations\footnote{\footnotesize The experiment videos are available at \url{https://www.qingniao-ai.cn/aeroagent}}.

\subsection{Wildfire Search and Rescue Scenario}\label{fire}

\subsubsection{Scenario Description}

The scenario of a wildfire search and rescue assesses the agent's ability to perform complex tasks. This scenario was selected for initial evaluation because it closely parallels real-world industrial drone applications: it entails diverse objectives, extended mission durations, and the presence of significant uncertain information. Additionally, potential accidents serve as a measure of AeroAgent's skills in contingency planning.

In this exercise, the drone embarks on its mission from a mountaintop starting point, positioned at some distance from the wildfire. Initially, the drone's vision of the fire and smoke is obscured, necessitating a closer approach for accurate assessment. Within the forest, there are four groups of trapped individuals totaling eight people (2, 1, 3, 2), as well as three firefighters. The drone is tasked with reducing environmental interference, relaying real-time updates to the command center, establishing verbal communication to reassure those trapped, and delivering emergency supplies. Given the mission's tight schedule and limited supply drops, precise execution is crucial to maximize mission efficacy.

While enhancing the task efficiency of AeroAgent through the input of more detailed objectives would be advantageous, the initial exclusion of such information underscores the reinforcement learning limitations in integrating a priori knowledge. Consequently, this constraint more aptly demonstrates AeroAgent's emergency response proficiency and its capacity for few-shot learning.

\subsubsection{Scenario Goal Design}

The Scenario Goal in reinforcement learning is understood as the reward. Therefore, the goals we designed are rewards that need to be explored in the environment for reinforcement learning. Whereas for AeroAgent, to ensure fairness, we also do not explicitly input these goals. However, experimental results show that AeroAgent can leverage its own harnessed world knowledge to reason well about what needs to be done in the current situation.

The goal structure for this scenario encompasses several key components, as listed below:

\begin{itemize}
\item Approaching the fire source: This provides strong guidance for reinforcement learning, which struggles to perform well without auxiliary rewards. Before getting within 3 meters of the fire source, the reward obtained is inversely proportional to the distance from the fire source, and is standardized so that the maximum reward obtained from this item is 10 points.
    
\item Reporting on the fire: Information relay regarding the fire garners a flat reward of 10 points.
    
\item Reporting on individuals who are trapped: Each confirmed report of a trapped person's situation increases the reward by 2 points, scoring no more than 16 points in total.
    
\item Consoling trapped individuals via communication: Each successful reassurance awards 2 points, with a ceiling of 16 points.
    
\item Distribution of emergency kits to those in need: Delivering an emergency kit to a trapped person contributes an additional 2 points to the overall reward, to a maximum of 16 points.
    
\item Misidentification penalties: Mistakenly categorizing firefighters as trapped individuals results in a 1-point deduction per occurrence.
\end{itemize}

\subsubsection{Perception and Action}

In our experiments involving LMMs, we preprocess the multimodal data obtained from onboard RGB and infrared sensors, encoding the imagery as Base64 strings via ROSchain or other approaches. For our reinforcement learning tasks, the sensory information undergoes transformation into multi-channel tensors.

To simplify RL task complexity, we introduce a set of discrete actions specifically designed for each scenario. This mitigates the challenge that arises from the vast continuum of potential commands available in ROSchain, such as move or turn, which feature an extensive parameter space.

We detail the discrete action space as follows:

\begin{itemize}
    \item Navigate to a designated location defined by proximity: close, medium, or distant.
    \item Gather and report intelligence on the ignition point.
    \item Evaluate and communicate the count of individuals requiring rescue, with options from one to three.
    \item Establish communication lines with individuals in distress.
    \item Allocate and dispatch emergency kits, with quantities adjusted to the immediate need with options from one to three.
    \item AeroAgent also has the option to activate sensors, but reinforcement learning involves the integration of information from all sensors.
\end{itemize}

\subsubsection{Experiment Results}

\begin{figure}[h]
    \centering    \includegraphics[width=\linewidth]{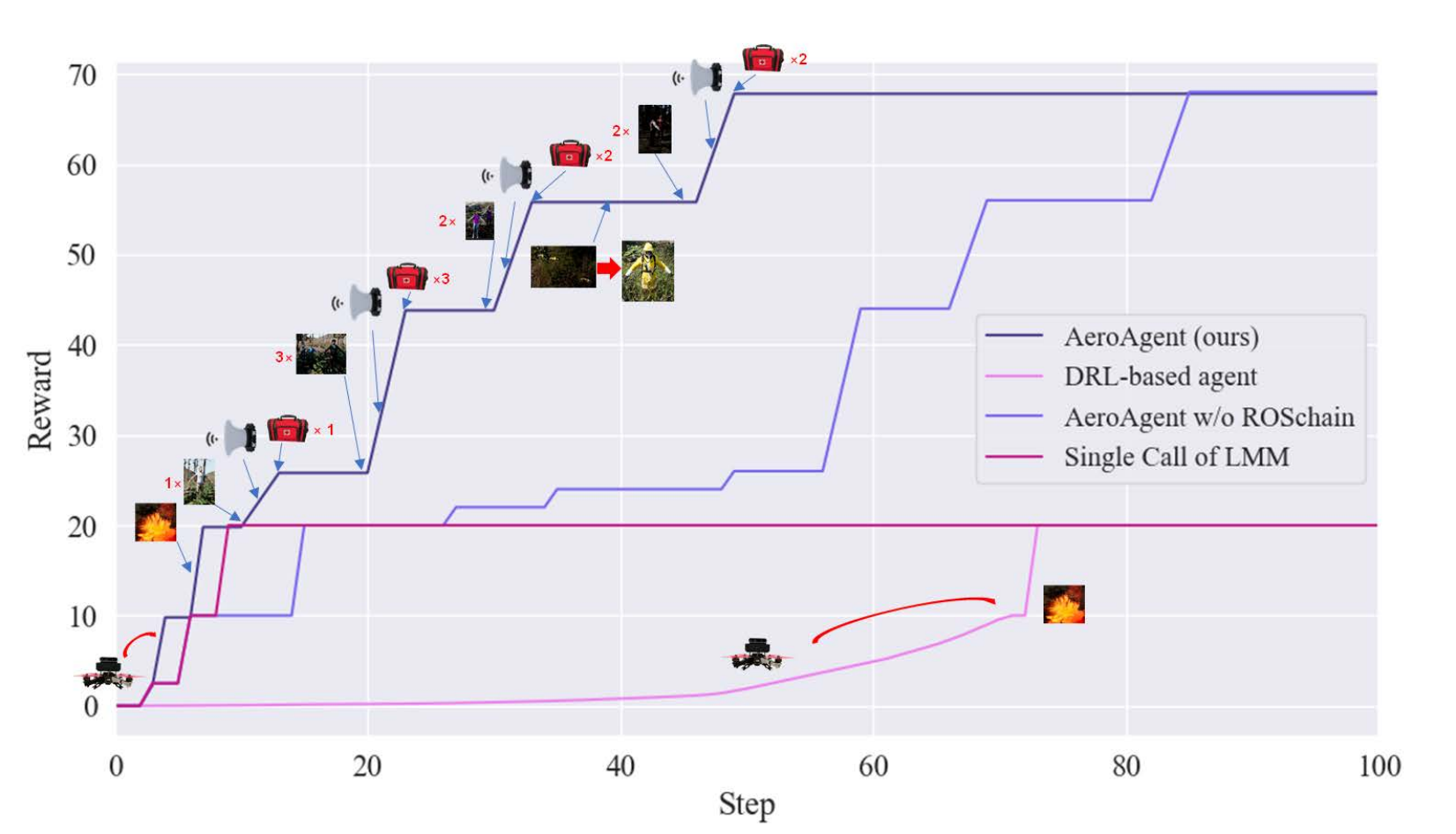}
    \caption{Results of wildfire search and rescue Scenario.}
    \label{fig:fire}
\end{figure}

We record the experimental process and results in Fig. \ref{fig:fire}.

The AeroAgent is designed to execute tasks with maximum efficiency and inherent speed optimization. In our ablation study conducted on ROSchain, we incorporated the Iterative Prompting Mechanism as discussed in Voyager\cite{wang2023voyager} to furnish the agent with a mechanism for feedback during instances where incorrect commands fail to elicit a response. This mechanism facilitated trial-and-error learning for the agent, specifically on how to interact efficiently with physical environments. However, our empirical analysis indicated that this approach exhibited significant latency compared to ROSchain's direct execution speed.

Moreover, we observed that although the Single Call of LMM can rapidly navigate to target locations using image comprehension, it falls short in retaining crucial planning and memory functionalities upon arrival—resulting in repetitive fixations on identified fire sources and a marked inability to formulate multi-stage operational tasks.

For the DRL-based Agent, it proves adept at reaching specified destinations guided by initial reward incentives and is competent at identifying fire hazards. However, it faces substantial challenges during the extended exploration stages. Even attempts to enhance performance by incorporating records of completed tasks into its feature set failed to generate improved outcomes. While advanced techniques in reinforcement learning may offer a solution to this setback, such solutions do not typically align with the more widely accepted generic design principles favored within the industry.




\subsection{Vision-based landing Scenario} 

\subsubsection{Scenario Description}

In the context of autonomous drone operations, the ability to accurately perform vision-based landings is of paramount importance, particularly in complex industrial environments. We present a simulation-based investigation focusing on drones executing precision landings on the helipads of unmanned offshore oil platforms. These helipads are marked with distinct visual cues to aid in the landing process. However, the presence of potential obstructions such as equipment debris and other drone traffic necessitates advanced visual recognition and precise control mechanisms to achieve high-accuracy landings. The system's proficiency in navigating to the designated landing point, guided solely by visual inputs, is a critical determinant of its operational success for autonomous takeoffs, landings, and cargo transportation within industrial contexts. Our study examines the efficacy of this approach and discusses the technological frameworks that enable such capabilities, contributing to safer and more reliable drone applications in offshore oil industry settings.

\subsubsection{Scenario Goal Design}

Our objective in this simulation is to orchestrate an optimal landing on the helipad, with the aim of touching down as close to the center point of the helipad as feasibly possible. In contrast to a prior scenario where auxiliary rewards were incorporated to alleviate the challenges inherent in reinforcement learning approaches, this current task will not factor in auxiliary rewards within the performance evaluation criteria. All other parameters remain consistent with the previously established scenario.

For this particular scenario, the goals incorporate the following items:

\begin{itemize}
    \item Executing a successful touchdown within the confines of the helipad boundary is incentivized with a reward of 10 points.
    \item The exactness of the landing's horizontal coordinate is critical: a precise landing at the helipad's center accrues a maximum of 10 points, whereas touchdown on the peripheral limit of the helipad garners no points. For landings falling within intermediate regions, the reward is calculated based on an inverse proportionality rule with respect to the distance from the helipad's center.
\end{itemize}

\subsubsection{Perception and Action}

Perceptions are the same as the previous scenario.

In this scenario, the fine maneuvering of the drone is very important for its landing spot. AeroAgent can focus on generating the movement coordinates for each step. For reinforcement learning, precise control requires a continuous action space. Therefore, we have designed the continuous action space for reinforcement learning as follows:

\begin{itemize}
    \item Taking into account the time and speed of each step, the action space is divided into three dimensions (x, y, z), with respective ranges of (-5, 5).
\end{itemize}

\subsubsection{Experiment Results}

\begin{figure}[h]
    \centering    
    \includegraphics[width=\linewidth]{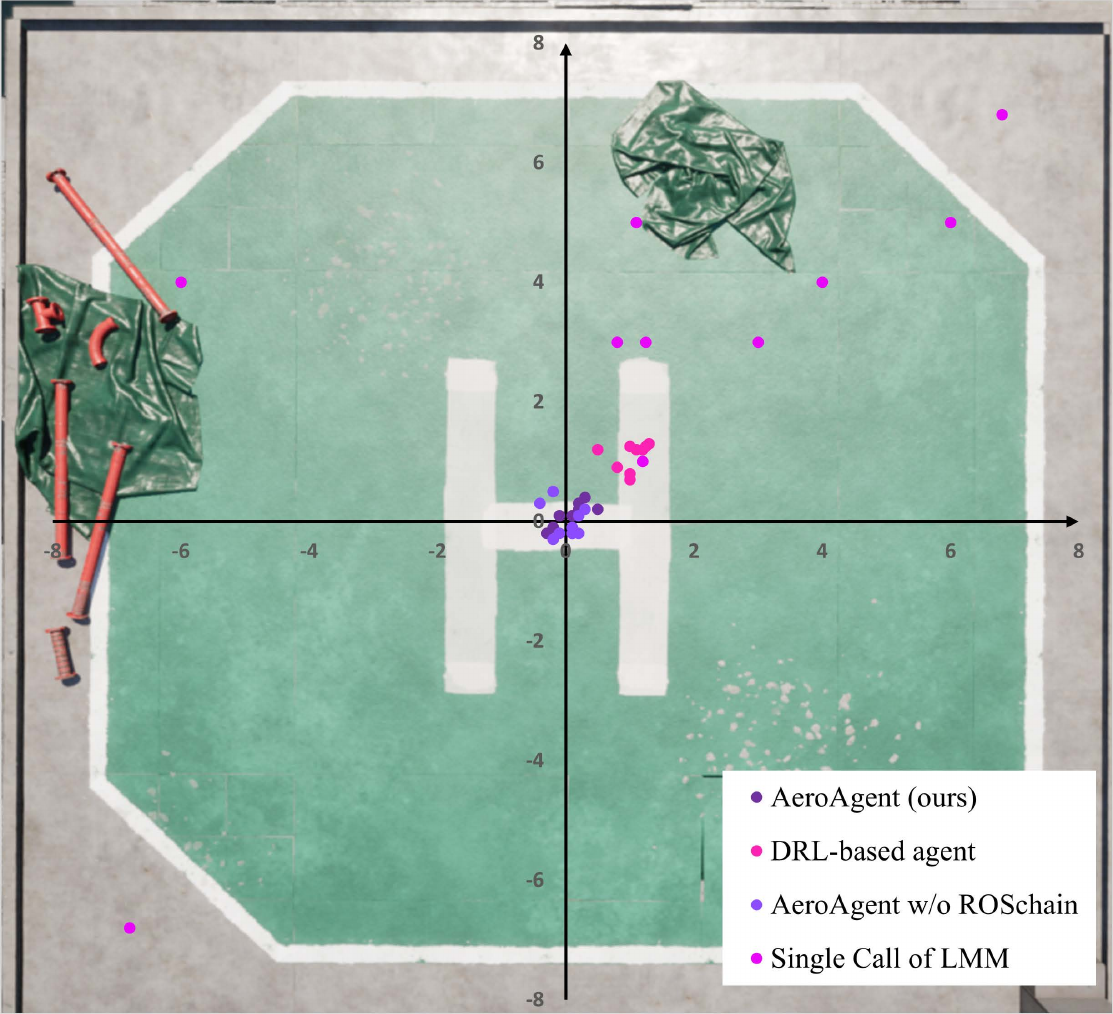}
    \caption{Results of Vision-based landing Scenario.}
    \label{fig:land}
\end{figure}

We have recorded the landing points of ten test results for each method, as shown in the Fig. \ref{fig:land}.

The AeroAgent demonstrates proficiency in task execution, notably achieving precise landings that consistently target the center of the apron. Comparative analysis between AeroAgent with and without the integration of ROSchain reveals near-identical performance metrics. However, we note that AeroAgent with ROSchain engages in a more complex initialization process, as evidenced by an increased number of steps executed during the commencement of movement directives. Furthermore, we observe that agents using deep reinforcement learning can be effectively trained to perform tasks with a high degree of competence. Nevertheless, these DRL-based agents have limitations in their exploration strategies, rendering them unable to reliably identify globally optimal solutions. Regarding the influence of sensor precision, the Single Call of LMM exhibits constraints in its exploratory capabilities, which manifests in fewer opportunities to rectify landing trajectories. This phenomenon, exacerbated by sensor inaccuracies, hinders the agent's ability to achieve consistently superior landing positions.

\subsection{Infrastructure Inspection Scenario}
\subsubsection{Scenario Description}

In scenarios involving infrastructure inspection, drones are deployed to detect faults in wind farm equipment amidst challenging conditions, such as high-altitude wind turbines and transmission lines, extensive transmission routes, and the hazards of direct contact with live wires. When contrasted with tasks like vision-based landing, which have been previously addressed using reinforcement learning, our experimental findings suggest that the complexities of infrastructure inspection make it exceedingly difficult for RL approaches to succeed. Specifically, these tasks often require the performance of subjective written assessments based on situational analysis, a process traditionally executed by human inspectors. Although drones can operate autonomously across a range of controls, the dynamic surveillance of exceptional circumstances necessitates a level of human subjective judgment that is reflected in the evaluation of images or videos transmitted from the drones. Consequently, we refrained from designing experiments employing RL for these tasks as the predefinition of reporting content as discrete actions for a DRL-based agent lacks practical viability in the context of industrial applications where the conditions and required actions are not sufficiently discrete or predictable.

\subsubsection{Scenario Goal Design}

In our study, we focus on the task of accurate fault detection in autonomous systems. Specifically, we predefine a fault scenario in which the right-side wind turbine ceases rotation. The evaluation metric for the agent's performance is a reward function that quantifies the accuracy of fault reporting.

\begin{itemize}
    \item An agent that does not report any fault will receive a zero reward. Conversely, when an agent reports a fault, we employ an LLM to ascertain the degree of correspondence between the reported information and the predefined fault. The rewards are structured on a ten-point scale, where a precise alignment with the defined fault condition yields the maximum reward of 10 points.
\end{itemize}

\begin{table*}[h]
\caption{Comparison of experimental results$^{\mathrm{a}}$}
\begin{center}
\begin{tabular}{c|c c c c c c c c}

\hline 
& \multicolumn{2}{c}{\textbf{\textit{Wildfire Search and Rescue }}}& \multicolumn{2}{c}{\textbf{\textit{Vision-based Landing }}}& \multicolumn{2}{c}{\textbf{\textit{Infrastructure inspection}}}& \multicolumn{2}{c}{\textbf{\textit{Safe navigation}}}
\\
\hline
&TR&AR$^{\mathrm{b}}$&TR&AR&TR&AR&TR&AR\\
\hline
Single Call of LMM & 29.4 &0.20&68.0&34.0& 0.0&0.0&11.1&0.46 \\

\hline

DRL-based Agent& 29.4& 0.20&93.8&4.69& N/A&N/A  &11.1&0.11\\

\hline

AeroAgent w/o ROSchain& \textbf{100.0}& 1.18&96.8&16.1 &\textbf{100.0}&11.1&\textbf{88.9}&3.70 \\

\hline
AeroAgent (ours)& \textbf{100.0} &\textbf{2.04}& \textbf{97.4} & \textbf{48.7}&\textbf{100.0}&\textbf{50.0}&\textbf{88.9}&\textbf{4.44}\\

\hline
\multicolumn{9}{l}{$^{\mathrm{a}}$Each reward is standardized to a maximum of 100 and a minimum of 0.}\\
\multicolumn{9}{l}{$^{\mathrm{b}}$TR = Total Rewards, AR = Average Step Rewards.
}
\end{tabular}
\label{tab1}
\end{center}
\end{table*}

\subsubsection{Perception and Action}

In the context presented, the perceptual inputs remain consistent with those established in the preceding scenarios. Given the impracticality of engaging a DRL-based agent for experimental purposes in this case, the associated action space for reinforcement learning does not merit consideration. With respect to the LMM action space, the primary requirement involves the coordination of movement commands with the respective report contents.

\subsubsection{Experiment Results}

\begin{figure}[h]
    \centering    
    \includegraphics[width=\linewidth]{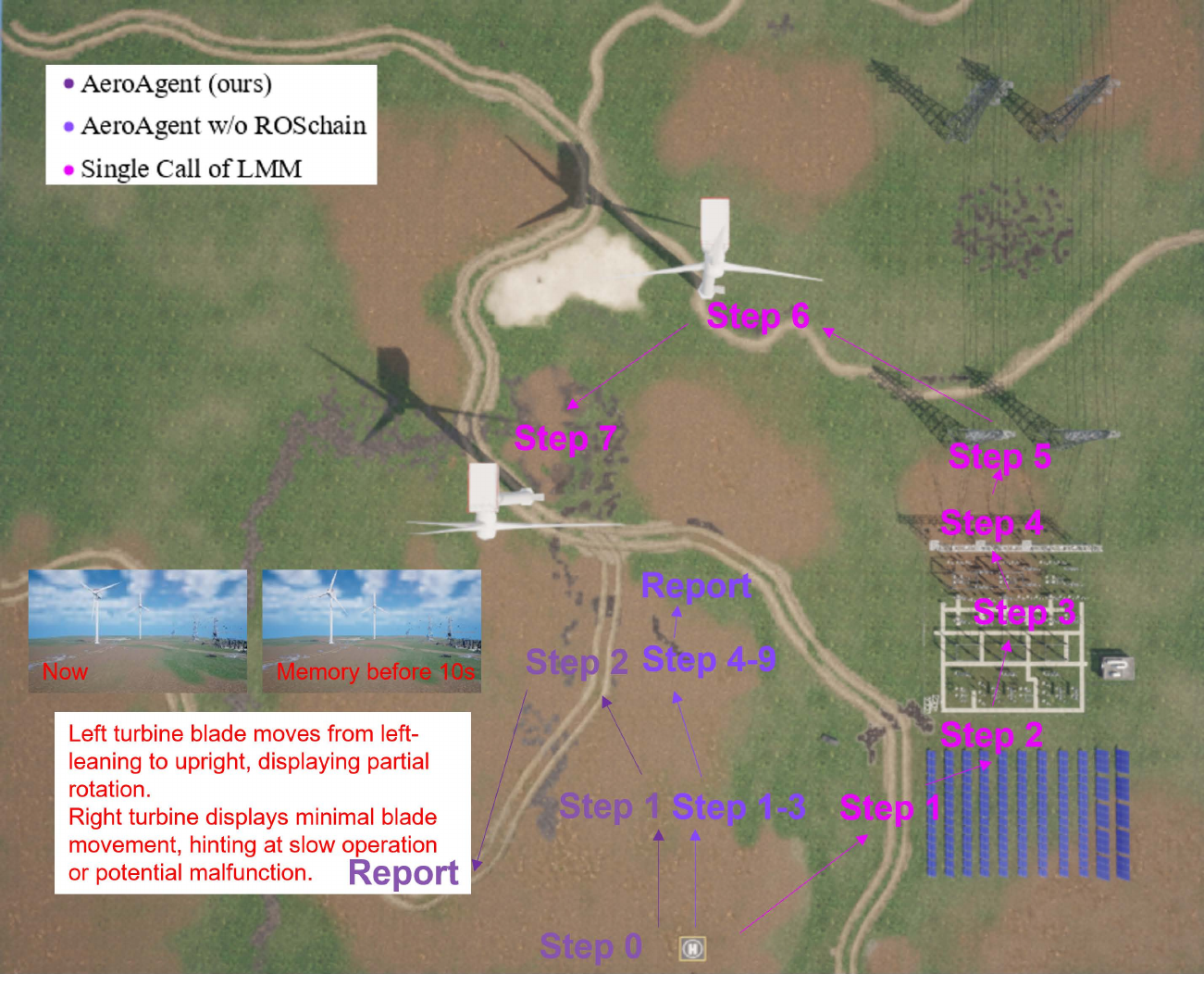}
    \caption{Results of Infrastructure Inspection Scenario.}
    \label{fig:power}
\end{figure}

We record the drone's movement path and key decision-making information during a test experiment, as illustrated in the Fig. \ref{fig:power}.

AeroAgent is endowed with memory functions that enable it to swiftly evaluate alterations in the environment by contrasting the latest image captured with preceding ones. As a result, it can detect abnormalities such as a normally rotating wind turbine on the left side, while identifying a malfunction in the motionless turbine on the right.

In contrast, AeroAgent without ROSchain displays comparable faculties but faces challenges in producing consistent commands during movement and error reporting, often necessitating repeated attempts informed by feedback.

Furthermore, a single call of LMM is deficient in memory components. It is restricted to processing only immediate environmental input, thereby proving inadequate at interpreting changes over time. Even after several trials, the LMM cannot recognize the fault in the stationary fan, ultimately falling short of fulfilling the assigned task.

\subsection{Safe Navigation Scenario}

\subsubsection{Scenario Description}

The concept of a safe navigation scenario encompasses a context in which a drone autonomously explores and patrols a complex environment, leveraging visual perception capabilities. This scenario is distinguished from conventional drone tasks such as path planning and obstacle avoidance directed at prespecified target locations. In the context of this study, the drone's mission is to maximize exploration of an abandoned factory space, methodically entering as many rooms as possible while concurrently ensuring its own safety through effective obstacle detection and avoidance. Crucially, the drone must independently evaluate its immediate surroundings to dynamically create waypoints for path planning and to strategically seek out feasible entry and exit points for continuous navigation. Traditional path generation and search techniques predicated on fixed start and end points are unsuitable in this setting, primarily because the drone's navigation is not guided by a predefined route, but rather by an adaptive exploration strategy void of explicit directional objectives.

\subsubsection{Scenario Goal Design}

In the proposed scenario, the objective is to maximize the number of building entries by an autonomous drone, which lacks any prior knowledge of the industrial complex's layout, thereby precluding extensive pre-mission planning. To mitigate challenges associated with sparse rewards that may hinder the drone's exploration process, our approach includes strategic reward shaping that ensures the proximity of rewards within the initial building.

\begin{itemize}
    \item It should be noted that the environment encompasses nine distinct buildings. The agent is designed to receive a reward of 10 points for each successful entry into any of these buildings.
\end{itemize}

\subsubsection{Perception and Action}

In this scenario, perceptions are the same as in previous scenarios. The action design is identical to that of vision-based landing.

\subsubsection{Experiment Results}

\begin{figure}[h]
    \centering    
    \includegraphics[width=\linewidth]{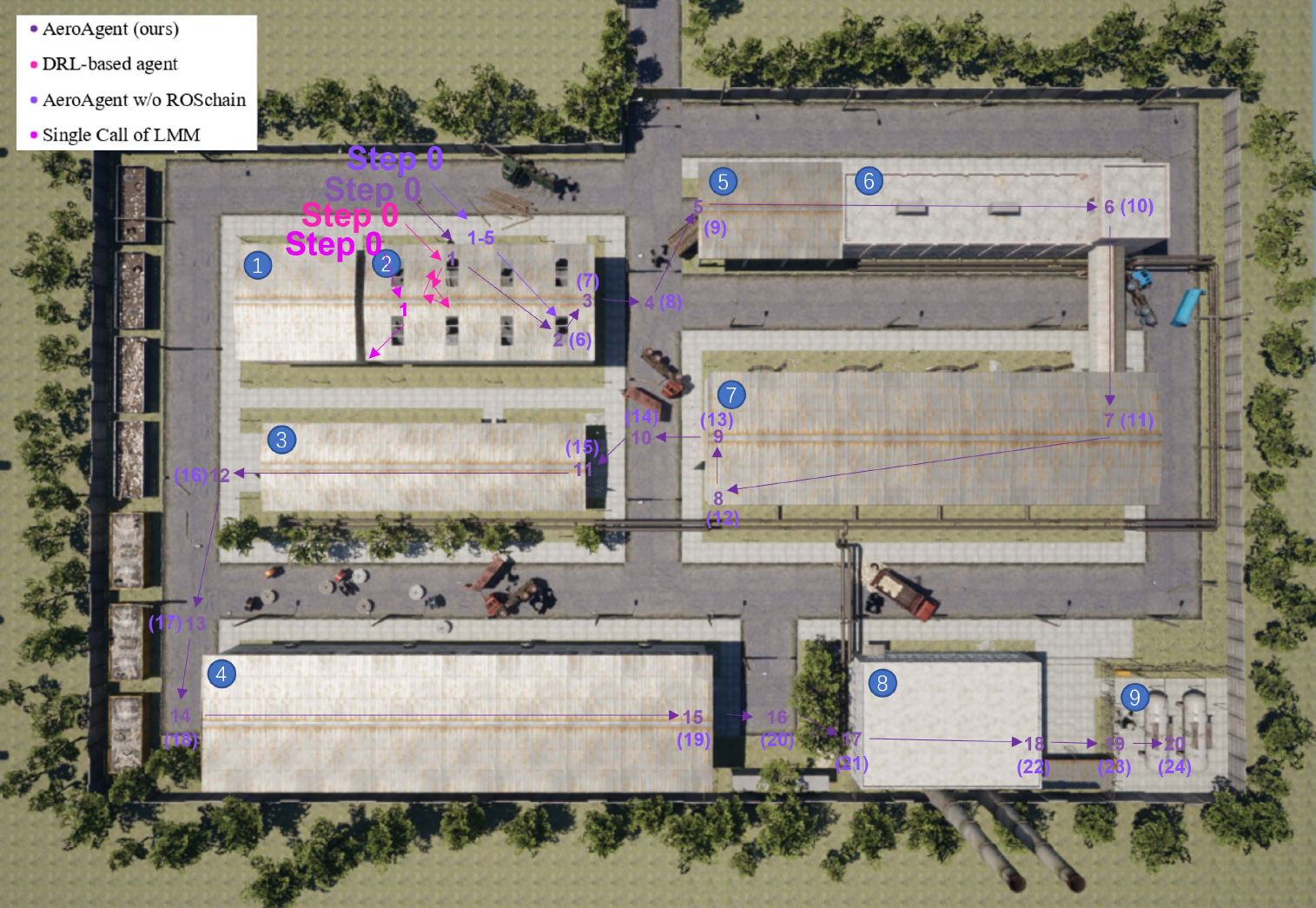}
    \caption{Results of Safe Navigation Scenario.}
    \label{fig:safe}
\end{figure}

We have recorded a drone's movement path during an experimental test, as depicted in the Fig. \ref{fig:safe}.

(1) Both AeroAgent and its counterpart lacking ROSchain capabilities were proficient in exploring various buildings. However, an exception was a small building located proximately to the initial area, which remained unexplored. Moreover, ROSchain exhibited proficiency in generating commands within a reduced timeframe and with greater stability.

(2) The agent powered by deep reinforcement learning shows a significant dependence on supplementary reward. To illustrate, such incentives were exclusively provided for navigation within the initial building, thereby impeding the exploration of subsequent buildings.

(3) In scenarios where the LMM is invoked for a single call, its absence of memory and plan components becomes evident. This limitation restricts the agents to merely processing real-time situational data. Consequently, the agents encounter difficulties when tasked with navigating out of intricate rooms due to insufficient data regarding potential egress points.

\section{Case Study}

In the preceding chapter, the comparative analysis of algorithmic performance between AeroAgent and DRL-based Agent was conducted within a simulated environment. Yet, a critical inquiry persists: What is the efficacy of AeroAgent's system architecture in executing practical industrial drone operations? Fundamentally, our principal contribution lies not in merely exceeding algorithmic performance benchmarks but in the cultivation of intelligent embodied agents capable of proficiently undertaking real-world tasks.

This chapter presents a case study—specifically, the individual search and rescue scenario—to critically evaluate the AeroAgent system's capacity for effective human-agent communication and strategic task planning.

\subsection{Case Design}

Drone-assisted rescue operations are a crucial application in the field of industrial drone use. As discussed in the experimental section of Chapter \ref{fire}, rescue drones frequently face sudden and unpredictable emergencies, which hinders pre-event data storage and scenario planning. Typically, rescue drones are manually operated for active exploration and situation monitoring. However, prompt response is vital in emergencies, and the real-time human operation of drones only extends the operator's sensory and action capacities. The true potential of drones in emergency management is realized when they independently monitor and manage critical situations. We previously assessed the AeroAgent's autonomous decision-making efficiency in simulations involving emergency scenarios. In this chapter, we describe an experiment conducted in an actual setting, where a drone equipped with AeroAgent autonomously guides a participant out of a simulated danger zone.

\begin{figure}[h]
    \centering    
    \includegraphics[width=\linewidth]{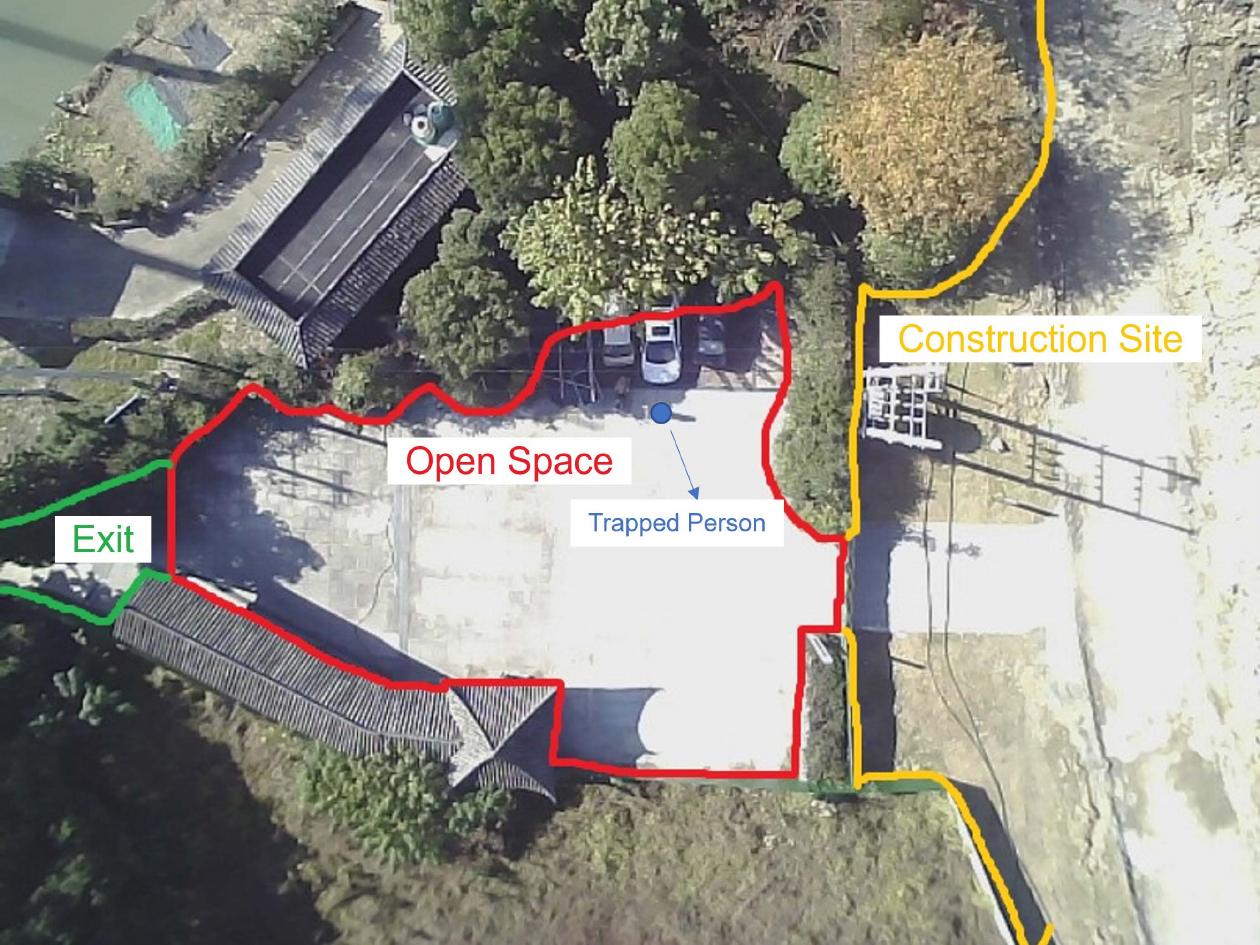}
    \caption{Individual search and rescue case study. The experimental results indicate that AeroAgent can effectively integrate various sensors and actuators to achieve environmental perception, analysis, decision-making, and execution.}
    \label{fig:design}
\end{figure}

For our case study, we chose an open space encircled by a construction site for an emergency simulation exercise. We envisaged a situation wherein an individual, while walking nearby, unintentionally wanders into the hazardous territory and becomes disoriented, as depicted in Fig. \ref{fig:design}. The path to the construction site gate on the open space's right is sealed, with barriers including walls, high-rise buildings, trees, and water bodies obscuring available exits from the person's line of sight.

During the experiment, a drone assigned to patrol the construction site is conducting its regular patrol and early warning tasks aloft. Its mission is to detect and warn of dangerous situations.

\subsection{Case Description}

\begin{figure*}[t]
    \centering    \includegraphics[width=\linewidth]{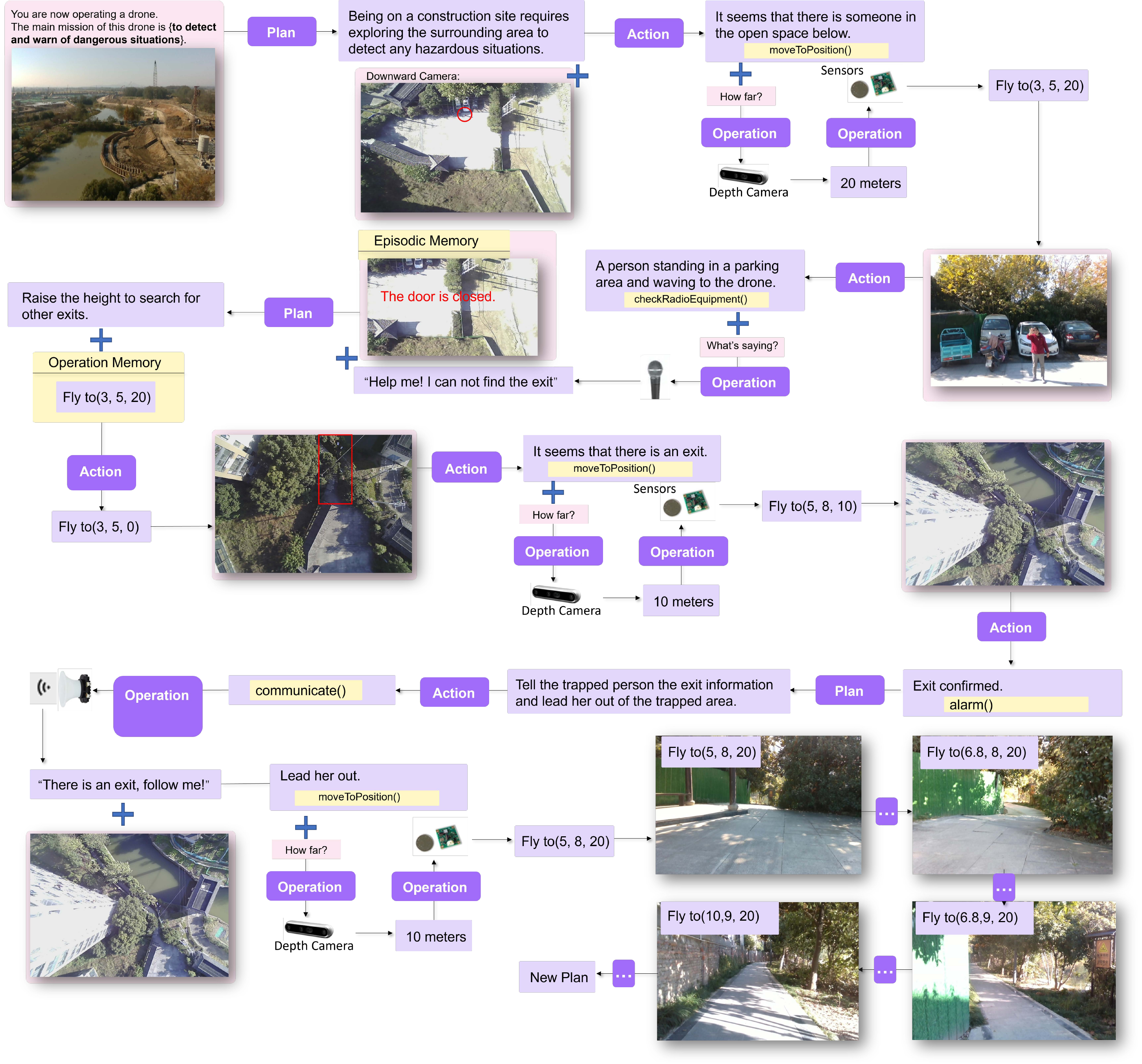}
    \caption{The case study process develops in the direction of the arrow. Firstly, it does not represent the preparatory phase, known as the standby process, which precedes the onset of surveillance tasks at construction sites. Secondly, the post-mission phase is absent from the illustration. Typically, a drone tasked with construction site supervision would resume patrolling if the power reserve was sufficient. 
    Thirdly, the diagram abstracts only the core procedures.}
    \label{fig:real}
\end{figure*}

In our case, AeroAgent functions autonomously, without the need for human intervention via remote control. 
Based on the experimental observations, the following insights have been garnered:

(1) Operative scenarios often present certain degrees of external interference, necessitating iterative outputs from the model. The robustness exhibited by AeroAgent in such conditions is commendable.

(2) The evaluations conducted did not encompass extended patrol operations, and notably, the mission's initiation was autonomous, not governed by a ground control station. AeroAgent autonomously performs situational analysis at predetermined intervals post-takeoff, effectively activating the mission upon detecting an individual in distress on the ground, thereby validating its standby efficacy.

(3) Within the system architecture, AeroAgent functions akin to an application layer on the drone platform. Subsequent to the dispatch of a movement directive to ROS, priority is managed by the drone’s flight control layer, which is integrated with an obstacle avoidance system that plays an instrumental role during navigational operations.


\section{Related Works}

\subsection{Robot Learning}

Self-supervised learning techniques, coupled with large-scale simulations, have been utilized to circumvent the need for resource-intensive supervised training in robot learning. The primary objective is to facilitate the transfer of skills acquired within simulated environments to real-world settings, as demonstrated by \cite{pinto2017learning}\cite{lynch2020learning}\cite{berscheid2019improving}\cite{yu2020meta}\cite{james2020rlbench}\cite{mittal2023orbit}\cite{zhu2020robosuite}\cite{tobin2017domain}\cite{shridhar2022cliport}\cite{handa2023dextreme}. Despite these advancements, research in multi-task reinforcement learning often remains constrained to simulated environments or exhibits a lack of task diversity when applied to real-world contexts, as noted in the works of \cite{espeholt2018impala}\cite{ song2019v}\cite{mandi2023cacti}. In the domain of imitation learning, the leveraging of expansive, real-world datasets has been a game-changer for enhancing generalization capabilities. This approach has been buoyed by the advent of substantial collections such as RT-1 and RT-2, as recently introduced by \cite{brohan2023rt1}, which hold remarkable promise for further advancing the field.

\subsection{Large Models in Robot}

Within this interdisciplinary domain, researchers often utilize methodologies such as Few-shot and Transfer Learning to adeptly tailor pre-trained models, including GPT-3 and BERT, for grounding activities with modest amounts of additional training. Notable constraints encompass an insufficient foundational comprehension, effectively addressed through approaches like fine-tuning and multimodal learning frameworks \cite{mahowald2023dissociating}\cite{yang2023llmgrounder}\cite{carta2023grounding}.

In the domain of Few-shot Planning, Large Language Models  are capable of producing decisions predicated on a limited set of examples, particularly evident in the processing of natural language inquiries to synthesize actionable plans and engage in zero-shot navigation tasks, a focal point of the study by \cite{zhou2023esc}.

When it comes to Zero-shot Navigation, LLMs excel by interpreting verbal instructions, formulating adaptive strategies, assimilating multimodal inputs, facilitating instantaneous interaction, and broadly generalizing across varied navigation challenges. Their applications prove crucial in the realms of directive processing, dynamic planning, and the aptitude for real-time adaptiveness in scenarios marked by novelty.

Two prevalent classes of Vision-Language Models (VLMs) are distinguishable. The first revolves around representation-learning frameworks like CLIP which seek to generate conjoint embeddings for visual and textual information. Conversely, the second category encompasses models capable of transforming vision and text inputs into elaborate textual output \cite{gan2022vision}\cite{radford2021learning}. Serving as a foundation for pre-training, these models significantly bolster the performance of tasks in object classification, detection, and segmentation \cite{gu2021open}. The investigation\cite{jinzhoulindevelopment} targets the latter variety of VLMs, which are particularly relevant to an array of functionalities, inclusive of image captioning and visual question answering. These models undergo training over a multiplicity of tasks utilizing extensive datasets.


\section{Conclusions, Limitations and Future Work}

 In this paper, we introduced a novel paradigm comprising an \verb|agent as cerebrum,| \verb|controller as cerebellum| framework for performing industrial tasks, an embodied agent architecture leveraging LMMs, and a ROSchain linkage framework connecting the agent to the ROS. Collectively, these components significantly contribute to advancing embodied agent technology for drone applications. We evaluated AeroAgent on the Airgen, where it demonstrated notable advantages in performance over existing DRL-based agents. Our ablation studies further elucidate the benefits contributed by ROSchain. In addition, case studies have confirmed the framework's utility within industrial contexts.

However, our work is not without its limitations. The real-world scenarios tested thus far were relatively simplistic and lacked the complexity of industry-level tasks. Consequently, our future endeavors will focus on extending the duration and complexity of tasks assessed, such as autonomous drone herding and automated power line inspections. These future experiments will also involve a broader array of actuators in order to thoroughly evaluate the embodied intelligence of our agents. Throughout these developments, we will continuously refine the AeroAgent architecture and ROSchain to ensure a high degree of stability and reliability.

\bibliographystyle{plain}
\bibliography{ref}



\end{document}